\documentclass[letterpaper, 10 pt, conference]{ieeeconf}
\usepackage{kotex}
\IEEEoverridecommandlockouts
\usepackage{subfigure}
\usepackage{amsmath}
\usepackage{cite}
\usepackage{bbold}
\usepackage{epsfig}
\usepackage{hyperref}

\usepackage{multirow}
\usepackage{multicol}
\usepackage{cellspace}
\usepackage{booktabs}

\usepackage{amssymb}
\usepackage{makecell} % For multi-line cells
\usepackage{graphicx} % For \resizebox
\usepackage{array}    % For \arraystretch
\usepackage{adjustbox} % for resizing
\def\thickhline{\noalign{\hrule height.8pt}}

\newcolumntype{P}[1]{>{\centering\arraybackslash}p{#1}}

\title{\LARGE \bf
Context-Based Visual-Language Place Recognition}

\author{Soojin Woo and Seong-Woo Kim
\thanks{All authors are with Seoul National University, Seoul, South Korea.\newline
 {\tt\footnotesize \{soojin.woo,snwoo\}@snu.ac.kr}}}
\begin{document}
\maketitle
\thispagestyle{empty}
\pagestyle{empty}

%%%%%%%%%%%%%%%%%%%%%%%%%%%%%%%%%%%%%%%%%%%%%%%%%%%%%%%%%%%%%%%%%%%%%%%%%%%%%%%%
\begin{abstract}
In vision-based robot localization and SLAM, Visual Place Recognition (VPR) is essential. This paper addresses the problem of VPR, which involves accurately recognizing the location corresponding to a given query image. A popular approach to vision-based place recognition relies on low-level visual features. Despite significant progress in recent years, place recognition based on low-level visual features is challenging when there are changes in scene appearance. To address this, end-to-end training approaches have been proposed to overcome the limitations of hand-crafted features. However, these approaches still fail under drastic changes and require large amounts of labeled data to train models, presenting a significant limitation. Methods that leverage high-level semantic information, such as objects or categories, have been proposed to handle variations in appearance. In this paper, we introduce a novel VPR approach that remains robust to scene changes and does not require additional training. Our method constructs semantic image descriptors by extracting pixel-level embeddings using a zero-shot, language-driven semantic segmentation model. We validate our approach in challenging place recognition scenarios using real-world public dataset. The experiments demonstrate that our method outperforms non-learned image representation techniques and off-the-shelf convolutional neural network (CNN) descriptors. Our code is available at \url{https://github.com/woo-soojin/context-based-vlpr}.
\end{abstract}

\section{Introduction}
Visual Place Recognition (VPR) is a crucial task in autonomous driving and robotics applications. In particular, in visual simultaneous localization and mapping (VSLAM), loop closure plays a vital role in correcting accumulated errors \cite{qian2022towards}. To achieve this, a robot must have the ability to determine whether its current location has been previously visited by comparing the incoming sensor data with a database during navigation. Thus, accurately recognizing the location corresponding to a given query image through VPR becomes an essential problem. Specifically, in the context of place recognition, it is important to recognize the same location even in large-scale environments where illumination may differ or where there are appearance changes over time. To address large environments, an approach that can compactly represent places while maintaining rich enough information to distinguish between visually similar locations is needed \cite{arandjelovic2016netvlad}.

One of the most prevalent methods used to tackle the place recognition problem in robotics is the visual bag-of-words (BoWs) approach, where local feature descriptors are used as visual words \cite{mur2014fast}. However, despite their success in many applications, camera-based VSLAM systems that rely on hand-crafted visual features such as SIFT \cite{lowe2004distinctive} and ORB \cite{rublee2011orb} are limited in that they cannot robustly handle changing environments because they rely on low-level features. These systems struggle particularly in scenarios involving changes in illumination and alterations to objects, such as their addition, removal, or rearrangement. \cite{mirjalili2023fm}. This limitation degrades the performance of loop closure detection (LCD), leading to distorted trajectory estimation and inaccurate map generation \cite{li2024resolving}.

Considering the shortcomings of hand-crafted features, recent approaches have proposed training convolutional neural network (CNN) networks in an end-to-end manner. End-to-end CNN models such as NetVLAD \cite{arandjelovic2016netvlad} outperform non-learned image representations and off-the-shelf CNN descriptors \cite{mirjalili2023fm}. Although these models demonstrate robustness to viewpoint and appearance variations, they still struggle with substantial changes in environment and objects. Additionally, a significant limitation is the need for labor-intensive dataset labeling for training. Thus, we focus on the use of high-level semantic information to address the problem of VPR in dynamic environments. % \cite{mirjalili2023fm}.

\begin{figure}[t]
    \vspace{0.2cm}
    \centering
    \framebox{\parbox{0.49\textwidth}{\includegraphics[width=0.485\textwidth, height=0.17\textwidth]{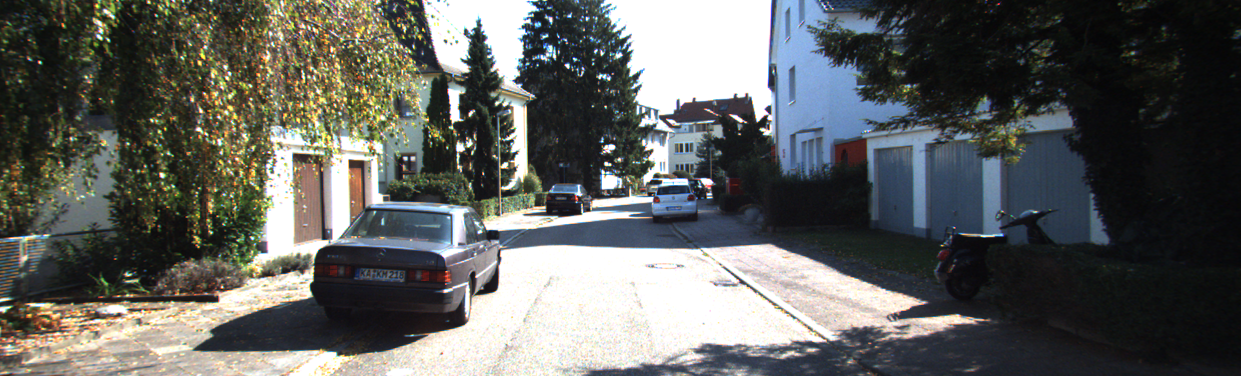}}}
    \vspace{0.2cm}
    
    \centering
    \framebox{\parbox{0.49\textwidth}{\includegraphics[width=0.485\textwidth, height=0.17\textwidth]{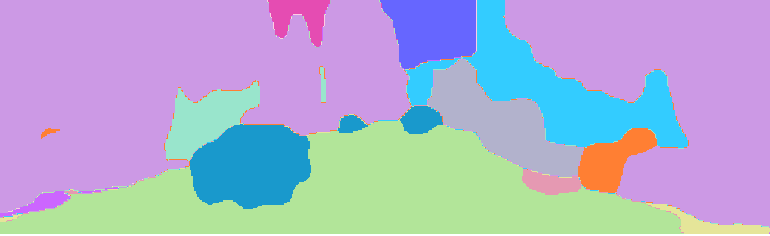}}}
    
    \caption{The language-driven semantic segmentation is based on a pre-defined label set. The segmentation results on the KITTI dataset were obtained by correlating per-pixel embeddings and text embeddings. The pre-defined labels include road, sidewalk, building, vehicle, car, bicycle, motorcycle, vegetation, trunk, terrain, cyclist, pole, sky, and other.}
    \vspace{-0.4cm}
    \label{fig:segmentation}
\end{figure}

In this paper, we propose a novel VPR method that operates robustly in dynamic scenes, based on a zero-shot, language-driven semantic segmentation approach \cite{li2022language}. The pre-trained language-driven semantic segmentation model is used to extract pixel-level language embedding information within the image. The extracted pixel-level language embedding information undergoes a processing stage to generate semantic image descriptors, which are then used to build a visual-language vocabulary in the visual BoWs place recognition module \cite{mur2015orb}. This approach enables place representation using a visual vocabulary that contains rich information extracted from the pixel-level semantic image descriptor. In addition, the image is represented as a single vector using BoWs, ensuring compactness. Consequently, this method efficiently addresses long-term VPR problems without relying on descriptors based on hand-crafted features (e.g., SIFT, SURF, ORB) \cite{stenborg2018long}. Furthermore, semantic segmentation assigns labels to each pixel corresponding to an object, allowing the distinction between static and dynamic objects based on preliminary knowledge \cite{kaneko2018mask}.

This capability not only enhances robustness in place recognition, even in changing environments but also enables handling new environments by easily modifying preliminary information. 

The main contributions of this paper are as follows:
\begin{itemize}
    \item \emph{Visual-language vocabulary-based place recognition system}: We introduce the concept of Visual-Language Vocabulary to generate a vocabulary using pixel-level semantic descriptors extracted from a large set of images. This vocabulary is then used to recognize the revisited locations.
    \item \emph{Context graph}: We propose the Context Graph concept, which helps understand the context within an image and provides robustness in complex environments. The nodes contain keypoint information, and the edges denote the distance between two different keypoints.
    \item \emph{Public release of code}: We have made the code used in our research publicly available, enabling other researchers to use it.    
\end{itemize}

\begin{figure*}[t]
\vspace{0.2cm}
    \centering
    \framebox{\parbox{\textwidth}{\includegraphics[width=\textwidth,height=6.0cm]{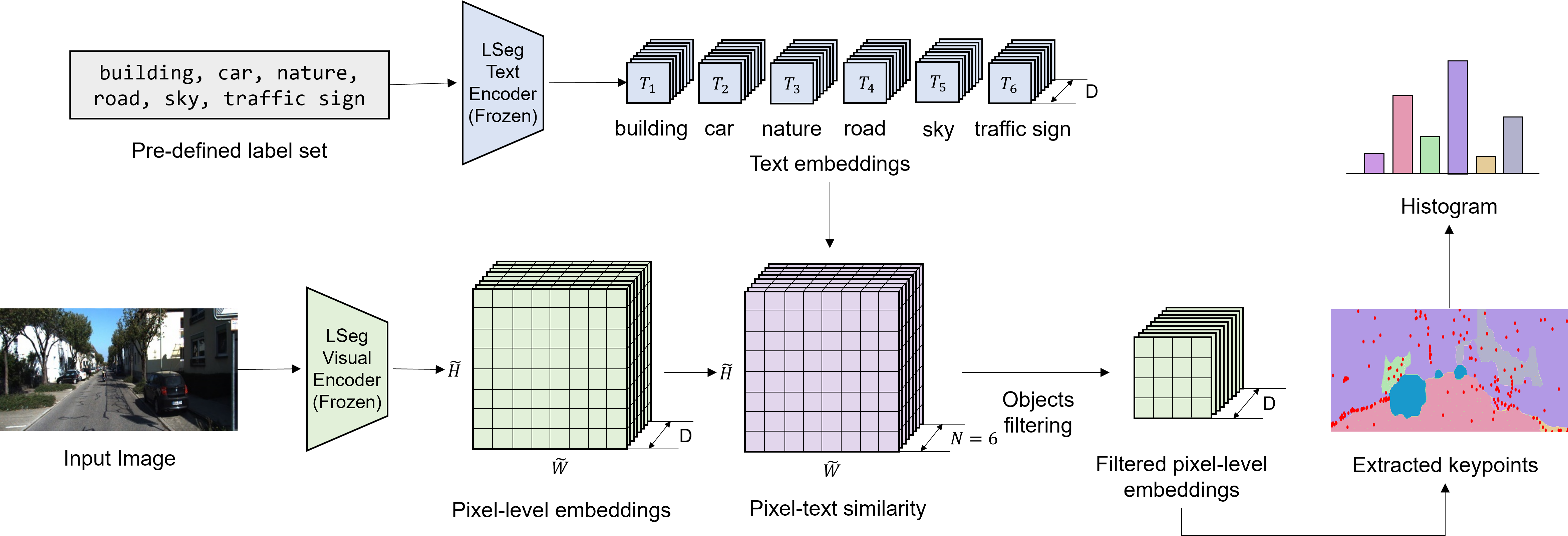}}}
    
    \caption{System Overview. The pre-trained text encoder of LSeg is used to generate text embeddings from a pre-defined label set. The visual encoder of LSeg extracts per-pixel embeddings from the image, which are then correlated with text embeddings to predict the class label for each pixel. The predicted labels are used to filter out pixel coordinates corresponding to dynamic objects, such as cars. After filtering, $K$ keypoints are randomly selected from the remaining pixel coordinates to create descriptors.}
    \label{fig:system_overview}
    \vspace{-0.2cm}
\end{figure*}

\renewcommand{\arraystretch}{2.0}
\begin{table}[t]
\centering
\resizebox{1.0\linewidth}{!}{
% \huge
\begin{tabular}{c|cccc}
Methods & \makecell{Illumination \\Change} & \makecell{Dynamic \\Environment} & \makecell{No \\Additional Training} & \makecell{Context \\Understanding} \\ \hline \hline
\makecell{Hand-crafted \\Feature-based} & & & \checkmark  & \\ \hline
End-to-end & \checkmark & &  & \\ \hline
Semantic & \checkmark &  \checkmark &  &  \\ \thickhline
\textbf{Ours} & \checkmark & \checkmark & \checkmark &  \checkmark\\ \hline
\end{tabular}
}
\caption{Comparison of Existing and Proposed Approaches.}
\label{table:table_comparison}
\vspace{-3.0em}
\end{table}
% \makecell{Long-term\\ Problem} 

\section{Related Works}
\subsection{Hand-crafted feature-based approaches}
To solve VPR and LCD, local feature-based methods have been introduced. One of the most prevalent approaches is the visual BoWs technique, which uses local feature descriptors as visual words. Among these methods, FAB-MAP \cite{cummins2008fab} integrates SURF and BoWs techniques for place recognition, demonstrating high efficiency and robust performance against viewpoint changes. DBoW2 \cite{galvez2012bags} introduced the highly efficient FAST feature detector \cite{rosten2006machine} combined with BRIEF descriptors \cite{calonder2010brief}, using bags of binary words. The use of BRIEF reduced the time required for feature extraction compared to traditional BoWs approaches that used SURF \cite{bay2006surf} and SIFT \cite{lowe2004distinctive}, resulting in a more efficient and robust system. However, it does not guarantee invariance to scale and orientation changes. To address this, ORB-SLAM \cite{mur2015orb}, a state-of-the-art method in Visual SLAM, proposed a place recognition approach based on ORB \cite{rublee2011orb} and BoWs methods, which are invariant to rotation and scale, enabling real-time LCD. Although it demonstrated high recall and robustness in experiments, hand-crafted features based on low-level information are susceptible to environmental changes, such as lighting variations.

\subsection{End-to-end approaches}
To overcome the limitations of hand-crafted feature-based methods, place recognition approaches using CNN architectures trained in an end-to-end manner have been proposed. The de facto standard architecture for end-to-end CNN models for place recognition is NetVLAD \cite{arandjelovic2016netvlad}. Patch NetVLAD \cite{hausler2021patch} enhances the basic NetVLAD model by obtaining patch-level features from NetVLAD residuals, enhancing its performance. Although these approaches demonstrate robustness to changes in appearance and viewpoint, they still struggle with severe variation in the environment and object arrangement. In addition, training-based methods require large amounts of labeled datasets, which involve labor-intensive labeling tasks. Recently, promising approaches have been proposed that utilize high-level semantic features to address dramatic appearance changes in LCD and VPR.

\subsection{Semantic approaches}
In Semantic SLAM, research has been conducted on the use of semantic information to perform loop-closure detection based on a more comprehensive understanding of the scene. Studies by Hu \emph{et al.} \cite{hu2019loop}, Li \emph{et al.} \cite{li2020view}, and Qian \emph{et al.} \cite{qian2022towards} use the relative positions of semantic objects to distinguish similar scenes for LCD. However, these methods are limited by not considering the characteristics of each object, making them less effective at distinguishing different places. Textslam \cite{li2020textslam} \cite{li2023textslam} employs text as planar features but is limited by its reliance on pixel-level features rather than on semantic information.

In contrast, Heikel and Espinosa-Leal \cite{heikel2022indoor} obtain object labels from the input image using YOLO \cite{redmon2016you} to perform a VPR task. These object labels are then converted into language-based features, which are used with a random forest classifier to predict room labels. FM-Loc \cite{mirjalili2023fm} addresses the challenge of VPR failures in changing environments by employing foundation models for object detection and scene classification. Specifically, it uses the large language model GPT-3 \cite{brown2020language} and the Visual-Language Model CLIP \cite{radford2021learning} to obtain semantic information. This semantic information, including object and place categories, is used to generate semantic image descriptors, which are then employed to compute the similarity between query and database images. Li \emph{et al.} \cite{li2024resolving} addresses the issue of existing low-level feature-based approaches that fail in seemingly similar environments by using semantic information. The authors use Yolov6 \cite{li2022yolov6} and East \cite{zhou2017east} for object and text detection, incorporating information such as the location of the boundaries, classes, and points of the map covered by the boxes into the semantic data. In addition, they employ Blip-2 \cite{li2023blip} and ChatGPT to generate textual descriptions for LCD.

The mentioned methods utilize semantic information about several objects within the image to perform localization. Consequently, they cannot capture the relationships between objects, or the information is insufficient to represent the entire image. As a result, these methods struggle when the distance threshold between the query and the database images is set too low. In addition, they are only applicable in specific scenarios, such as repetitive environments. Therefore, this paper proposes a method that works in more general scenarios, which will be introduced in the following sections. Table \ref{table:table_comparison} presents a comparison between the existing methods and the proposed approach.

In this paper, we propose a novel VPR methodology that leverages high-level semantic information to achieve robustness against environmental changes. Additionally, our approach does not require additional training by utilizing pre-trained weights and enables an understanding of image context by considering the relationship between segmentation regions.

\begin{figure*}[t]   \centering\includegraphics[width=\textwidth,height=0.3\textwidth]{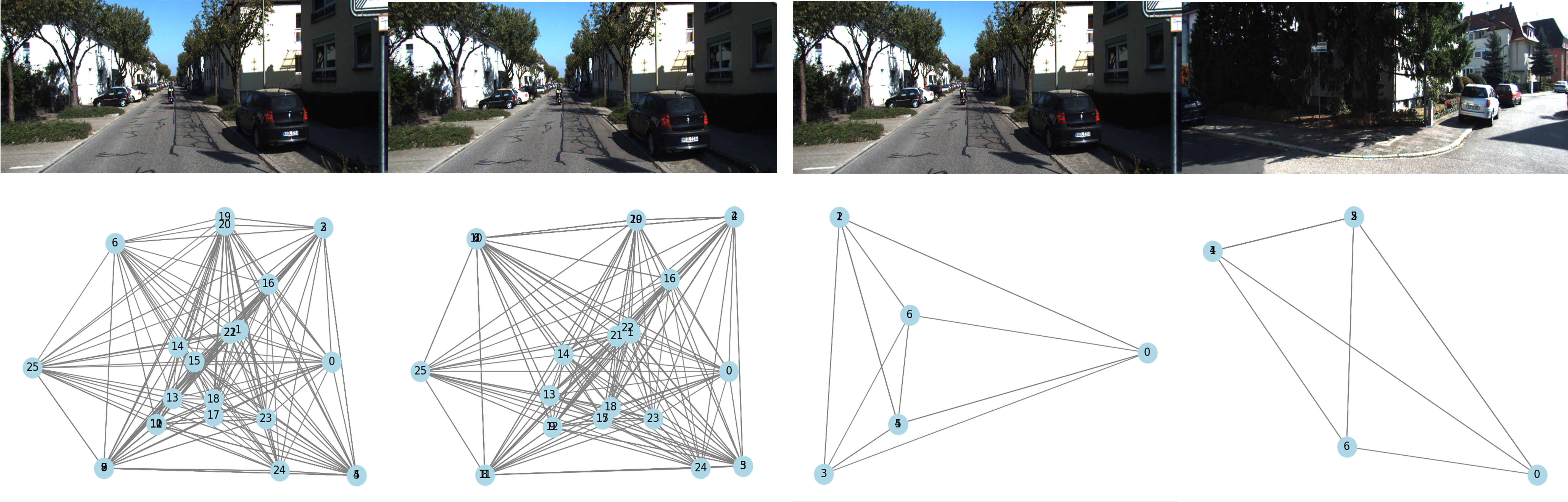}
    \caption{Context Graph. Each circle represents the centroid of a cluster and serves as a node in the context graph. The graph visualizes the context graph of each image. These are qualitative results showing that the context graph demonstrates greater similarity among visually similar images.}
    \vspace{-0.2cm}
    \label{fig:context_graph}
\end{figure*}

\section{Methods}
In this section, we describe our scene context-based visual-language place recognition approach. The process is described in Fig. \ref{fig:system_overview}. A set of labels including predefined categories for objects was created based on prior knowledge of the environment in which the robot operates. The set of labels was defined using the class information provided by the public dataset known as Semantic KITTI \cite{behley2019semantickitti}. After defining labels that contain information about objects, text embeddings for each label were extracted using the frozen text encoder of LSeg. Next, pixel-level embeddings are extracted from the input RGB image using the pre-trained visual encoder of LSeg. Thus, the correlation between text and pixel-level embedding vectors is calculated for a single image to determine the object category of each pixel. Figure \ref{fig:segmentation} shows the results of the pixel-level segmentation for the KITTI dataset based on a predefined label set. Subsequently, filtering is applied to pixels corresponding to dynamic objects, such as cars, which degrade VPR performance. The visual-language vocabulary is generated from the remaining pixel-level embeddings. This vocabulary is used to represent the entire image as a single histogram, which allows efficient retrieval of the candidate image from the database based on Euclidean distance.

This paper aims to achieve successful place recognition even in complex environments by using high-level semantic information. To this end, we propose a methodology that incorporates pixel-level semantic information while also considering the relationships between objects to understand the context of the image. The following subsections will look at the technical details of scene context-based visual-language place recognition, including (i) visual-language embedding, (ii) visual-language vocabulary, and (iii) context graph.

\subsection{Visual-Language Embedding}\label{methods:visual_language_embedding}
We use the visual-language model LSeg \cite{li2022language} to obtain pixel-level embedding information from image frames captured by the robot's camera. LSeg is a model designed for language-driven zero-shot semantic image segmentation, capable of segmenting RGB images based on free-form language categories. The text encoder embeds the given label set into a vector space, extracting embedding vectors. Based on prior knowledge of the robot's surrounding environment, we have predefined a set of labels for object categories. We calculate embeddings for these $N$ descriptive input labels, resulting in $N$ vectors $T_{1},...,T_{N} \in \mathbb{R}^{D}$. The size of an input image is assumed to be $H \times W$, while the output is downsampled to an image of size $\frac{H}{s} \times \frac{W}{s}$ using a downsampling factor $s$. The downsampled image size is defined as $\Tilde{H} = \frac{H}{s}$ and $\Tilde{W} = \frac{W}{s}$. Subsequently, a transformer-based image encoder calculates dense per-pixel embeddings, resulting in an output embedding $I \in \mathbb{R}^{\Tilde{H} \times \Tilde{W} \times D}$. To obtain segmentation information for each pixel $(i,j)$, the inner product between the image embedding vector $I_{ij}$ and the text embedding vectors $T_{1},...,T_{N}$ is computed. The correlation results in a vector $F_{ij} \in \mathbb{R}^{N}$ representing the predicted category for each pixel. The label $k$ corresponding to the maximum correlation $f_{ijk}$ among the $N$ language categories indicates the category of the object for the pixel. This can be mathematically expressed as follows:
\begin{gather}
    f_{ijk} = I_{ij} \cdot T_{k},\\
    F_{ij} = (f_{ij1}, f_{ij2}, ..., f_{ijN})^{T}.
\end{gather}

\subsubsection{Dynamic Objects Filtering}\label{methods:dynamic_objects_filtering}
From the previous step, we obtained the predicted category on the image through segmentation. Using the segmentation results, we can filter out dynamic objects that could potentially degrade VPR performance by predefining such categories.

If the label of the pixel $(i,j)$ matches a predefined category, the embedding vector $I_{ij}$ corresponding to that pixel is filtered and not used when generating the descriptors. If the label of the pixel $(i,j)$ matches a predefined category, the embedding vector $I_{ij}$ corresponding to that pixel is filtered out so that it is not included in the descriptors. In the paper, object labels such as vehicle, car, bicycle, motorcycle, cyclist, and other were designated as filtering targets. There have been several approaches to remove potentially dynamic objects, such as parked cars, in the map building and update process \cite{pomerleau2014long, hur2015precise, woo2024no} for autonomous navigation \cite{kim2017autonomous}. However, all of these methods are based on low-level pixel- or point-based approaches, whereas our method operates in a more abstract and generalized manner.

\subsection{Visual-Language Vocabulary}\label{methods:bag_of_words}
We use the BoWs method based on visual-language descriptors to search for candidate images similar to the query image. For BoWs-based place recognition, we generate a visual-language vocabulary using embedding information corresponding to keypoints extracted from Sec. \ref{methods:visual_language_embedding}. The extracted keypoints exclude the coordinates of the pixels associated with dynamic objects, as detailed in Sec. \ref{methods:dynamic_objects_filtering}. A fixed number of $K$ keypoints is randomly extracted from each region. The pixel coordinates corresponding to these extracted keypoints are used to create the visual-language vocabulary by forming descriptors. The histogram of codewords from the generated vocabulary allows for a compact representation of the entire image as a single vector. The upper right part of Fig. \ref{fig:system_overview} shows the result of the visual-language vocabulary of the input image.

\subsection{Context Graph}
In SLAM, it is essential to identify the most similar image among the candidate images to the query image to reduce the accumulated errors through LCD. In this section, we introduce the context graph, which considers both segmentation results and their relationships within the image, enabling VPR to understand the image context. Fig. \ref{fig:context_graph} illustrates the visualization of the context graph, and we will explain its details as follows.

\begin{figure}[t]
    \centering
    \subfigure[ORB Feature]{\includegraphics[width=0.485\textwidth, height=0.17\textwidth]{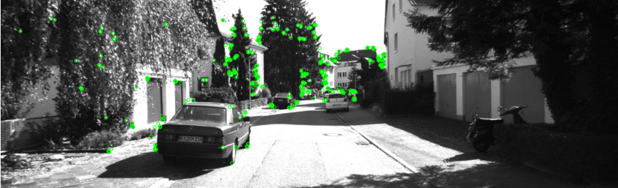}}\\[-1.0ex]
    \vspace{0.3em}
    \subfigure[Ours]{\includegraphics[width=0.485\textwidth, height=0.17\textwidth]{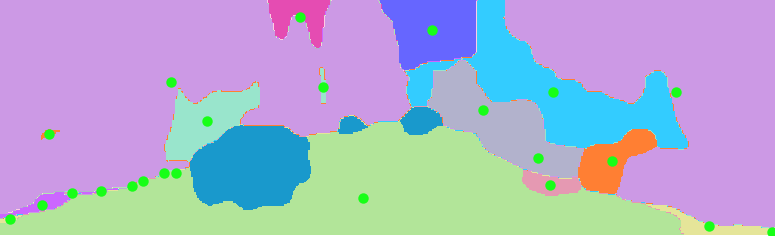}}\\[-1.1ex]
    \caption{Feature Extraction. (a) In the case of using ORB features, keypoints are extracted even from pixels corresponding to dynamic objects such as cars. In addition, very few features are extracted from the right side of the image, leading to uneven feature extraction across the entire image. (b) In our approach, filtering is applied so that features are not extracted from cars, and features are evenly extracted across the entire image. Additionally, our method uses fewer features compared to ORB, demonstrating an advantage in terms of computing efficiency.}
    \label{fig:feature_extraction} \vspace{-1.3em}
\end{figure}

To perform the correspondence matching between two images, the keypoints must be extracted. For this purpose, we first cluster pixels with the same label among the $N$ predefined object categories and calculate the centroids of each cluster to serve as keypoints. Then, the extracted keypoints are used as nodes in the context graph. Fig. \ref{fig:feature_extraction} illustrates the difference between the prior approach and ours, where our approach filters out dynamic objects, such as cars, that can degrade the performance of VPR. To find correspondences between the extracted keypoints, we first check whether each keypoint in the query and database images belongs to the same object category. If this condition is satisfied, we compute the cosine similarity between the pixel-level embeddings of the corresponding pixel coordinates. If the cosine similarity between the two embedding vectors exceeds a threshold $t$, which is set to $t=0.95$ in this paper, they are considered a valid correspondence. The pixel coordinates corresponding to these valid correspondences are then regarded as nodes when constructing the context graph. The approach can be described mathematically as follows:
\begin{gather}
   CosineSimilarity(I_{q}, I_{db_{i}}) = 1 - \frac{I_{q} \cdot I_{db_{i}}}
   {\Vert I_{q} \Vert \Vert I_{db_{i}} \Vert}.
\end{gather}
In the context graph $G=(V,E)$, the node set $v_{ij} \in V$ includes the coordinates of the keypoints. The edge set $e_{(ij,kl)} \in E$ stores the Euclidean distance $d$ between pixels, and an edge is created if $d \leq \tau$. The query image and the candidate images are represented by context graphs: $G_{Q} = (V_{Q},E_{Q})$ for the query image and $G_{D} = (V_{D},E_{D})$ for the candidate images. The adjacency matrix $A_{G}$ is used to compare the context graphs of different images. It can be mathematically formulated as follows:
\begin{gather}
   d(I_{ij}, I_{kl}) = \sqrt{(x_i - x_k)^2 + (y_j - y_l)^2},\\
   e_{ij,kl} = d(I_{ij}, I_{kl}),\\
   V_{Q} = \{v_1,v_2,...,v_{n_{Q}}\},\\
   V_{D} = \{u_1,u_2,...,u_{n_{D}}\},\\
   V_{Q} = \{(x_i,y_i)|(x_i,y_i) \in Q\},\\
   V_{D} = \{(x_j',y_j')|(x_j',y_j') \in D\},
\end{gather}
\begin{gather}
    (A_G)_{ij} =
    \begin{cases}
    1 & \text{if } d(v_i, v_j) \leq \tau, \\
    0 & \text{otherwise}
    \end{cases}.
\end{gather}
The context graph generated for the query image $G_{q}$ is compared with the context graphs of the candidate images $G_{db_{i}}$, and the candidate image with the most similar context graph to $G_{q}$ is ultimately selected.

\begin{table*}[ht]
\renewcommand{\arraystretch}{1.5}
\centering
\begin{tabular}{l||cccc|cccc}
\toprule
\textbf{Approach} & \multicolumn{4}{c|}{NetVLAD\cite{arandjelovic2016netvlad}} & \multicolumn{4}{c}{Proposed} \\
& r@1 & r@5 & r@10 & r@20 & r@1 & r@5 & r@10 & r@20\\
\midrule
Pitts30k-val\cite{torii2013visual} & 0.54 & 0.91 & 1.00 & 1.00 & 0.32 & 0.76 & 0.96 & \textbf{1.00} \\
KITTI00\cite{geiger2012we} & 0.12 & 0.26 & 0.35 & 0.46 & 0.10 & \textbf{0.37} & \textbf{0.52} & \textbf{0.66}\\
\bottomrule
\end{tabular}
\caption{Performance comparison of different layers.}
\label{table:recall_comparison}
\end{table*}

\section{Experiments}
In this section, we describe the datasets used to validate our approach in Sec. \ref{experiments:evaluation_dataset}, the evaluation metrics in Sec. \ref{experiments:evaluation_metrics}, the quantitative results in Sec. \ref{quantitative_analysis} and the qualitative results in Sec. \ref{qualitative_analysis}.

\subsection{Datasets}\label{experiments:evaluation_dataset}
To evaluate our methodology, we used two publicly available datasets: Pittsburgh \cite{torii2013visual} and KITTI \cite{geiger2012we}. These datasets contain numerous dynamic objects, such as cars and pedestrians, as well as changes in illumination and viewpoint. They were chosen to demonstrate the robustness of our approach in dynamic environments.

\textbf{Pittsburgh.} The dataset generated from Google Street View includes test queries captured at different times, spanning several years. In our study, we used the Pitts30k dataset for evaluation. In the Pitts30k-val dataset, 10,000 database images and 7,608 query images are provided. For the experiments, a subset of this dataset was randomly sampled, and the evaluation code was made publicly available. 

\textbf{KITTI.} The dataset was acquired using a stereo camera mounted on a moving vehicle and includes real-world image data captured from urban, rural, and motorway scenes. For our study, we utilized the left color images from the KITTI dataset. Since the dataset was originally used for odometry, it includes data from additional sensors such as GPS. To evaluate our method, we constructed the database by considering only the 2D coordinates from the 3D GPS ground truth poses. We trained a k-nearest neighbors (kNN) model using the coordinates of the database data and calculated distances to query images to find the nearest neighbors within a distance threshold (e.g., $d=25$ meters). For the evaluation, all 4,541 images from sequence 00 were utilized. Among these, 2,271 images were used as queries, and the remaining 2,270 images served as the database.

\subsection{Evaluation metrics}\label{experiments:evaluation_metrics}
We follow standard place recognition evaluation methods. A query image is considered accurately localized when at least one of the top $N$ database images returned by the proposed method is within $d = 25$ meters of the query's ground truth position, with the distance calculated using Euclidean distance. The percentage of correctly localized queries is computed based on the top $N$ database images. In this paper, we calculated recall for $N = 1, 5, 10, 20$. To perform VPR, we compute histograms for the query image $I_{q}$ and the database images $I_{db}$ and then calculate the Euclidean distance between these histograms. The database images that are closest in distance to the query image are selected as candidates. It can be formulated as follows:
\begin{gather}
   Recall = TP / (TP + FN), \\
   distance(I_{q}, I_{db_{i}}) = \sqrt{\sum(I_{q}-I_{db_{i}})^{2}}.
\end{gather}
\subsection{Visual Place Recognition (VPR)}
We evaluated the performance of VPR through experiments. Specifically, we assessed whether the VPR system could still identify previously visited locations even after a period of time, despite changes in the scene, using both quantitative and qualitative experiments.

\subsubsection{Quantitative evaluation}\label{quantitative_analysis}
We compared our method with the state-of-the-art appearance-based localization approach, NetVLAD \cite{arandjelovic2016netvlad}. Table \ref{table:recall_comparison} shows the results of the recall on Pittsburgh and KITTI. In the Pittsburgh dataset, our methodology demonstrated comparable performance to the NetVLAD trained on the Pittsburgh dataset. Specifically, on the KITTI dataset, our approach outperformed NetVLAD in terms of r@5, r@10, and r@20. Notably, while NetVLAD uses a vector size of 32,768 for the VPR task, our methodology represents images with a smaller vector, confirming that our approach achieves superior place recognition performance based on a more compact representation.

\subsubsection{Qualitative evaluation}\label{qualitative_analysis}
\begin{figure}[t]
    \centering
    \subfigure[ORB Feature]{\includegraphics[width=0.485\textwidth, height=0.15\textwidth]{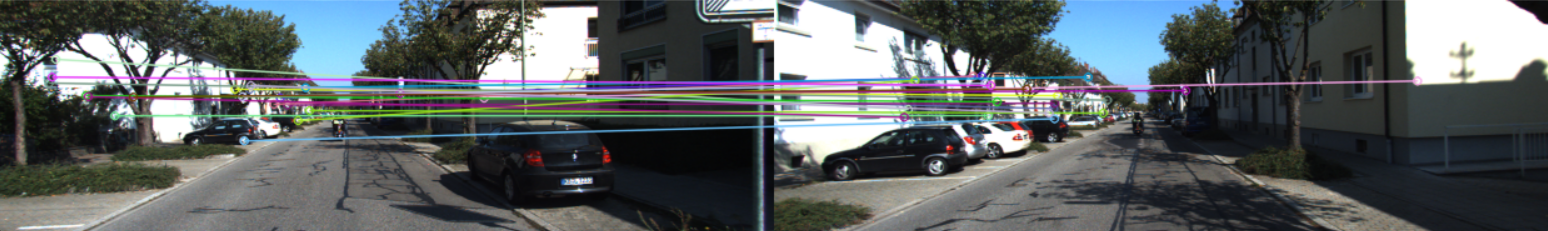}}\\[-1.0ex]
    \vspace{0.3em}
    \subfigure[Ours]{\includegraphics[width=0.485\textwidth, height=0.15\textwidth]{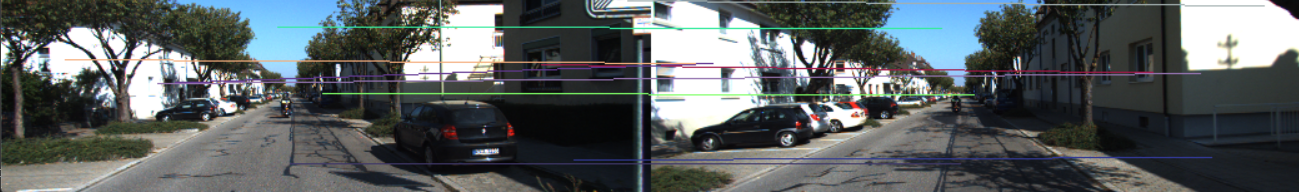}}\\[-1.1ex]
    \caption{Correspondence Matching. The results of correspondence matching are visualized as follows: (a) matching results based on ORB features and (b) matching results based on our method.}
    \label{fig:correspondence} \vspace{-1.3em}
\end{figure}

The qualitative evaluation of the correspondence matching results is shown in Fig. \ref{fig:correspondence}, with a comparison to existing hand-crafted feature extraction methods. By performing object filtering as detailed in Sec. \ref{methods:dynamic_objects_filtering}, we confirm that correspondence matching performs well in environments with numerous dynamic objects (e.g., cars). Additionally, utilizing dense pixel-level embedding information for image representation demonstrates robust performance even in environments with significant viewpoint and illumination changes. Furthermore, a qualitative comparison of image similarity based on the context graph is presented in Fig. \ref{fig:context_graph}, demonstrating the potential of VPR to understand the relationships between objects within the images. The proposed method outperforms traditional hand-crafted feature-based correspondence matching approaches.

\section{Conclusion}
In this paper, we proposed a novel scene context-based visual-language place recognition method that operates robustly in dynamic environments. By utilizing high-level semantic information, our approach overcomes the limitations of traditional low-level feature-based methods that struggle with severe environmental changes. Additionally, we conducted experiments on real-world datasets to demonstrate the robustness of our method. Since our method employs a zero-shot semantic segmentation model, it not only eliminates the need for inefficient labeling tasks but also makes it applicable in new environments without additional training. We utilized a visual-language model to extract pixel-level semantic descriptors, representing the entire image as a single vector to compare query and reference images. By applying the BoWs method, we ensured a compact representation of large-scale images while maintaining the richness through pixel-level embedding vectors. Furthermore, we incorporated preliminary knowledge to filter out pixels corresponding to dynamic objects, such as cars and pedestrians, validating through experiments that our method operates robustly even in dynamic scenarios. In addition, we propose a method for successful loop detection in robotics using the context graph. We calculate the similarity score between the query and candidate images to ultimately select the most similar one.

\section*{Acknowledgement}
This research was funded by the Korean Ministry of Land, Infrastructure and Transport (MOLIT) through the Smart City Innovative Talent Education Program and by the Korea Institute for Advancement of Technology (KIAT) under a MOTIE grant (P0020536). Research facilities were provided by the Institute of Engineering Research at Seoul National University.

\bibliographystyle{IEEEtran}
\bibliography{IEEEabrv, main}
\end{document}